\documentclass[letterpaper, 10 pt, conference]{ieeeconf}  % Comment this line out if you need a4paper

\IEEEoverridecommandlockouts                              % This command is only needed if 
\usepackage{graphics} % for pdf, bitmapped graphics files
\usepackage{epsfig} % for postscript graphics files
\usepackage{mathptmx} % assumes new font selection scheme installed
\usepackage{times} % assumes new font selection scheme installed
\usepackage{amsmath} % assumes amsmath package installed
\usepackage{amssymb}  % assumes amsmath package installed

\usepackage[above]{placeins}
\usepackage{subcaption} %Make the TABLE label same line as the text and prevent all text capitalized.
\usepackage[hyphens]{url}
\usepackage{xcolor}
\usepackage{xspace}
\usepackage{multirow}
\usepackage{pgfplots}
\definecolor{darkblue}{rgb}{0.196,0.314,0.784}
\usepackage[colorlinks=true]{hyperref}
\newcommand{\fig}[1]{Fig.~{#1}}

\newcommand{\comment}[1]{{}}

\newcommand{\RBDispNetC}{FADNetS\xspace}
\newcommand{\RBDispNetMC}{ESNet\xspace}
\newcommand{\RBDispNetMOC}{ESNet-M\xspace}

\makeatletter
\newcommand\notsotiny{\@setfontsize\notsotiny\@vipt\@viipt}
\makeatother
\title{\LARGE \bf
ES-Net: An Efficient Stereo Matching Network
}

\author{Zhengyu Huang$^{1}$, Theodore B. Norris$^{1}$ and Panqu Wang$^{2}$% <-this % stops a space
\thanks{*This work was done during Zhengyu Huang's internship at TuSimple.}% <-this % stops a space
\thanks{$^{1}$Zhengyu Huang and Theodore B. Norris are with the Department of Electrical and Computer Engineering, University of Michigan, Ann Arbor, MI 48105, USA.
        {\tt\small zyhuang@umich.edu; tnorris@umich.edu}}%
\thanks{$^{2}$Panqu Wang is with TuSimple, Inc.,
        San Diego, CA 92122, USA.
        {\tt\small panqu.wang@tusimple.ai}}%
}

\begin{document}

\maketitle
\thispagestyle{empty}
\pagestyle{empty}

%%%%%%%%%%%%%%%%%%%%%%%%%%%%%%%%%%%%%%%%%%%%%%%%%%%%%%%%%%%%%%%%%%%%%%%%%%%%%%%%
\begin{abstract}

Dense stereo matching with deep neural networks is of great interest to the research community. Existing stereo matching networks typically use slow and computationally expensive 3D convolutions to improve the performance, which is not friendly to real-world applications such as autonomous driving. In this paper, we propose the Efficient Stereo Network (\RBDispNetMC), which achieves high performance and efficient inference at the same time. \RBDispNetMC relies only on 2D convolution and computes multi-scale cost volume efficiently using a warping-based method to improve the performance in regions with fine-details. In addition, we address the matching ambiguity issue in the occluded region by proposing \RBDispNetMOC, a variant of \RBDispNetMC that additionally estimates an occlusion mask without supervision. \comment{We focus on efficient 2D convolution based network structure and study the effect of several stereo matching components on its performance, including multi-scale cost volume with warping and occlusion modeling.} We further improve the network performance by proposing a new training scheme that includes dataset scheduling and unsupervised pre-training. %On KITTI 2015 benchmark,% 
Compared with other low-cost dense stereo depth estimation methods, our proposed approach achieves state-of-the-art performance on the Scene Flow~\cite{SceneflowAndDispNetC}, DrivingStereo~\cite{DrivingStereo}, and KITTI-2015 dataset~\cite{KITTI2015}. Our code will be made available.

\end{abstract}

%%%%%%%%%%%%%%%%%%%%%%%%%%%%%%%%%%%%%%%%%%%%%%%%%%%%%%%%%%%%%%%%%%%%%%%%%%%%%%%%
\section{INTRODUCTION}
Depth estimation is one of the most fundamental problems in computer vision, which has a broad range of applications in robotics, virtual reality, and autonomous driving. Depth estimation can be achieved using monocular camera~\cite{eigen2014depth}\cite{MonoDepth}, stereo camera~\cite{SGM2007}\cite{LeCun2015}, light field camera~\cite{Lightfield_depth_Wang_2015}\cite{SPO_zhang2016}, and LiDAR point cloud. Among these methods, stereo camera based depth estimation is particularly attractive due to its advantage of low cost, high resolution and long working range. 

Stereo depth estimation is generally formulated as a matching problem: given a rectified stereo pair, depth can be estimated by identifying the correspondence between the pixels in the left-view image and the right-view image. Specifically, the depth $z$ of a pixel $(x,y)$ in the left-view image, with corresponding pixel in the right-view image at $(x-d,y)$, is given by $z = f\cdot b/d$, where $f$ is the focal length of the camera, $b$ is the separation between the left and right camera (baseline), and $d$ is referred to as disparity. Following the taxonomy of stereo matching~\cite{Stereo_Taxonomy}, traditional stereo matching can be broken down into four steps: (1) matching cost computation, (2) cost aggregation,  (3) optimization, and (4) disparity refinement. Based on the approach of optimization, stereo depth estimation can be categorized into local methods and global methods. Local methods perform a local winner-take-all disparity optimization at each pixel~\cite{veksler2003fast}\cite{yoon2006adaptive}, while global methods perform a global optimization of an energy function that includes both a data term (matching cost) and a smoothness term~\cite{woodford2009global}.

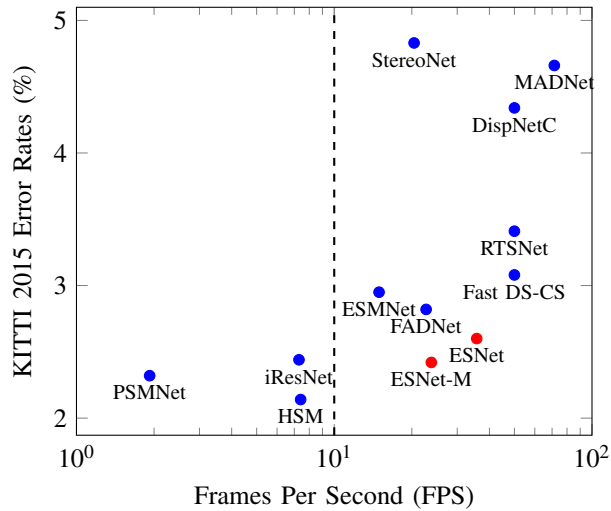
\begin{figure}[tb!]

\centering
\begin{tikzpicture}

\begin{axis}[xmode=log, xlabel= Frames Per Second (FPS), ylabel = KITTI 2015 Error Rates (\%), xmin=1, xmax=100,xlabel near ticks, ylabel near ticks]
    \addplot[
        scatter/classes={a={blue}, b={red}},
        scatter, mark=*, only marks,
        scatter src=explicit symbolic,
        nodes near coords*={\Label},
        nodes near coords align={north},
        every node near coord/.append style={font=\footnotesize},
        visualization depends on={value \thisrow{label} \as \Label} %<- added value
    ] 
    table [x expr=1/\thisrowno{0},y index=1, meta=class] {
        x y class label
         0.028 2.60 b \RBDispNetMC
         0.042 2.42 b \RBDispNetMOC
         %0.044 1.99 b $\text{FADNet}^*$
         0.044 2.82 a FADNet
         0.020 4.34 a DispNetC
         0.049 4.83 a StereoNet
         0.520 2.32 a PSMNet 
         1.030 2.87 a GC-Net

         0.137 2.44 a iResNet
         0.014 4.66 a MADNet 
         0.020 3.41 a RTSNet
         0.067 2.95 a ESMNet
         0.020 3.08 a Fast\space DS-CS
         0.135 2.14 a HSM
    };
    \draw[dashed, thick] (axis cs: 10,\pgfkeysvalueof{/pgfplots/ymin}) -- 
                      (axis cs: 10,\pgfkeysvalueof{/pgfplots/ymax});
\end{axis}
\end{tikzpicture}
\caption{KITTI 2015 Benchmark error rates versus FPS. Our proposed models (red) have the best speed/accuracy trade-off.} 
\label{fig:Timing Plot}
\vspace{-0.5cm}
\end{figure}

With the rise of the deep learning era, many works apply neural networks to stereo depth estimation. In a seminal work by Žbontar and Lecun~\cite{LeCun2015}, a CNN is trained to compare image patches and to compute the matching cost. While the cost is computed using learning-based method, the final disparity estimation is done in a classical way. Later works focus more on the end-to-end learning of stereo depth estimation: Mayer et al. propose the \mbox{DispNet-C}~\cite{SceneflowAndDispNetC} where the network architecture is an encoder-decoder structure like U-Net~\cite{Unet} and contains a novel correlation layer to compute the similarity scores between the features extracted from the left-view and the right-view images. Following this line, CRL~\cite{CRL} and FADNet~\cite{FADNet} improve the performance by cascading multiple networks to refine the disparity estimation. Besides the 2D network architectures, networks with 3D convolution layers are also used. Representative methods in this category include GC-Net~\cite{GCNet} and PSMNet~\cite{PSMNet}. These 3D convolution based methods preserve more information in the cost volume (4D) and typically lead to better performance compared to 2D methods. However, these 3D convolution based methods have high memory requirements and long inference time, making them inapplicable for real-time applications such as robotics and autonomous driving.

To this end, developing an efficient stereo matching network with good performance remains to be a challenge. In this paper, we propose a novel network, named as Efficient Stereo Network (\RBDispNetMC), to address this challenge. First, as stereo matching tends to have large error in fine-detailed regions, we propose a multi-scale feature matching module that computes the cost volumes at different scales efficiently using warping-based method and achieves higher accuracy. In addition, since stereo matching in occluded regions is ambiguous, we propose \RBDispNetMOC, a variant of \RBDispNetMC that is designed to learn an occlusion mask while estimating disparity. We show that such occlusion modeling improves disparity estimation accuracy. Finally, we also propose an unsupervised pre-training method and a dataset scheduling schema for stereo network training, which improve the accuracy without changing the inference time. By integrating the improvements we made above, we achieve the state-of-the-art performance compared to other low-cost dense stereo depth estimation methods on three public datasets, including Scene Flow~\cite{SceneflowAndDispNetC}, DrivingStereo~\cite{DrivingStereo}, and KITTI-2015 ~\cite{KITTI2015} (as shown in Fig.~\ref{fig:Comparison_KITTI}).

\section{RELATED WORK}
%\subsection{Efficient stereo matching networks}
\textbf{Efficient stereo matching networks} Most stereo matching networks can be classified as either based on 2D convolution or 3D convolution.\comment{A key component among 2D convolution based methods is the correlation layer. It was initially proposed in~\cite{FlowNet} for optical flow estimation and was later applied to stereo matching~\cite{SceneflowAndDispNetC}\cite{FlowNet2}.} Those 3D convolution based methods~\cite{GCNet}\cite{PSMNet} tend to have higher accuracy but sacrifice the speed, while 2D convolution based methods~\cite{SceneflowAndDispNetC}\cite{LeCun2015} have shorter inference time but larger error. To obtain a fast and accurate model, StereoNet~\cite{StereoNet} applies 3D convolution on a low resolution cost volume, without degrading the performance much. CRL~\cite{CRL} and FADNet~\cite{FADNet} both use a two-stage network design with a second network used to refine the disparity estimation from the first network. Although significant progress has been made, a simple network with low latency and high accuracy is yet to be developed.

\textbf{Multi-scale cost volume}
In the problem of optical flow estimation, to model large object motion, a multi-scale cost volume is constructed from the multi-scale feature maps, as in PWC-Net~\cite{PWCNet}. To compute the high resolution cost volume without significantly increasing the computation complexity, a warping operation is applied to the feature map of the target frame using the estimated coarse optical flow such that the possible motion magnitude between the source and target frame is reduced. This warping and pyramid-like structure leads to a very efficient algorithm, but is known to introduce ghosting effect in the occluded region~\cite{MFN} and cannot model small objects with large motion~\cite{Devon}. In addition, errors in estimated flow at low resolution will propagate to high resolution through warping and it is hard to correct them completely in the latter part of the network. To mitigate these limitations, Lu et al.~\cite{Devon} propose a dilated cost volume without warping, and Teed et al.~\cite{RAFT} propose a multi-scale cost volume without warping by pooling. In this paper, we adopt the idea from the PWC-Net to stereo depth estimation. Given the fact that potential errors could be introduced by warping at low resolution, our proposed \RBDispNetMC constructs multi-scale cost volume at higher resolution, which leads to improved performance.
%\cite{AANet}:construct multi-scale cost volume for stereo

\textbf{Occlusion modeling} 
Occlusion happens when a pixel in one view has no corresponding pixel in the other view. For these occluded pixels, the matching problem is undefined. Hence a proper occlusion modeling strategy is needed to obtain a good estimate of disparity in these occluded regions. \comment{However, the problem occlusion estimation are inherently coupled with the problem of depth estimation or more generally optical flow estimation.} Ilg et al.~\cite{Ilg2018_Occlusion} propose a network structure based on FlowNet2~\cite{FlowNet2} to jointly estimate optical flow (or disparity) and occlusion  with ground truth occlusion supervision. UnFlow~\cite{UnFlow} learns optical flow and occlusion mask by estimating a bidirectional flow, and occlusion is estimated by forward-backward consistency.\comment{\cite{Wang2018Occlusion_Flow}:Unsupervised learning of optical flow, with occlusion mask generated by backward flow and for warping.} Recently, Zhao et al.~\cite{MFN} propose an optical flow network (MaskFlowNet) with an asymmetric occlusion-aware feature matching module, which is able to estimate an occlusion mask without occlusion supervision. In our proposed \RBDispNetMOC, we handle the occlusion by an approach similar to MaskFlowNet and show its advantage on stereo matching performance.

\textbf{Unsupervised learning in depth estimation} 
Garg \mbox{et al.}~\cite{Garg2016unsupervisedMono} propose a CNN based unsupervised monocular depth estimation method, which only requires stereo image pairs for training. The network estimates a disparity map for an input left-view image and warps the right-view image using the estimated disparity to reconstruct the left-view image. The network is trained end-to-end by minimizing the photometric reprojection loss. Later work in MonoDepth~\cite{MonoDepth} achieves better performance by exploiting the left-right consistency.\comment{MonoDepth2~\cite{MonoDepth2} proposed the full resolution multi-scale photometric loss, per-pixel minimum reprojection and auto-masking to improve the model performance.} \comment{Training a network without supervision is useful when the training labels are hard to obtain. Unsupervised pre-training has been shown to be beneficial for tasks such as image classification~\cite{Unsupervised_pretrain}, optical flow~\cite{UnFlow}, object detection. However, it remains unclear how useful it is for the task of stereo matching. To our best knowledge, there is no prior work applying unsupervised learning for stereo matching network.}In this paper, we propose to pre-train stereo matching network in an unsupervised manner using the photometric reprojection loss and show significant improvement in the network performance.

\section{METHOD}

Since our goal is to achieve efficient stereo matching, we use FADNet's~\cite{FADNet} first network in its cascade as our baseline model for the study. We name this network \RBDispNetC, which is a 2D convolution based network with its structure similar to DispNetC ~\cite{SceneflowAndDispNetC}. Wang et al. ~\cite{FADNet} show that replacing the convolution layers used in the feature extractor of DispNetC by residual blocks~\cite{ResNet} leads to better accuracy in disparity estimation. However, \RBDispNetC only has a single scale cost volume and does not consider occlusion. This leaves opportunities for us to improve the design of the baseline network and to obtain a more accurate model with similar inference time (see discussion in section \ref{subsec:Key components}). \comment{We then describe in \ref{subsec:Loss} the loss function we used for supervised network training.}

\begin{figure*}[tb!]
\centering
\includegraphics[width=0.88\textwidth]{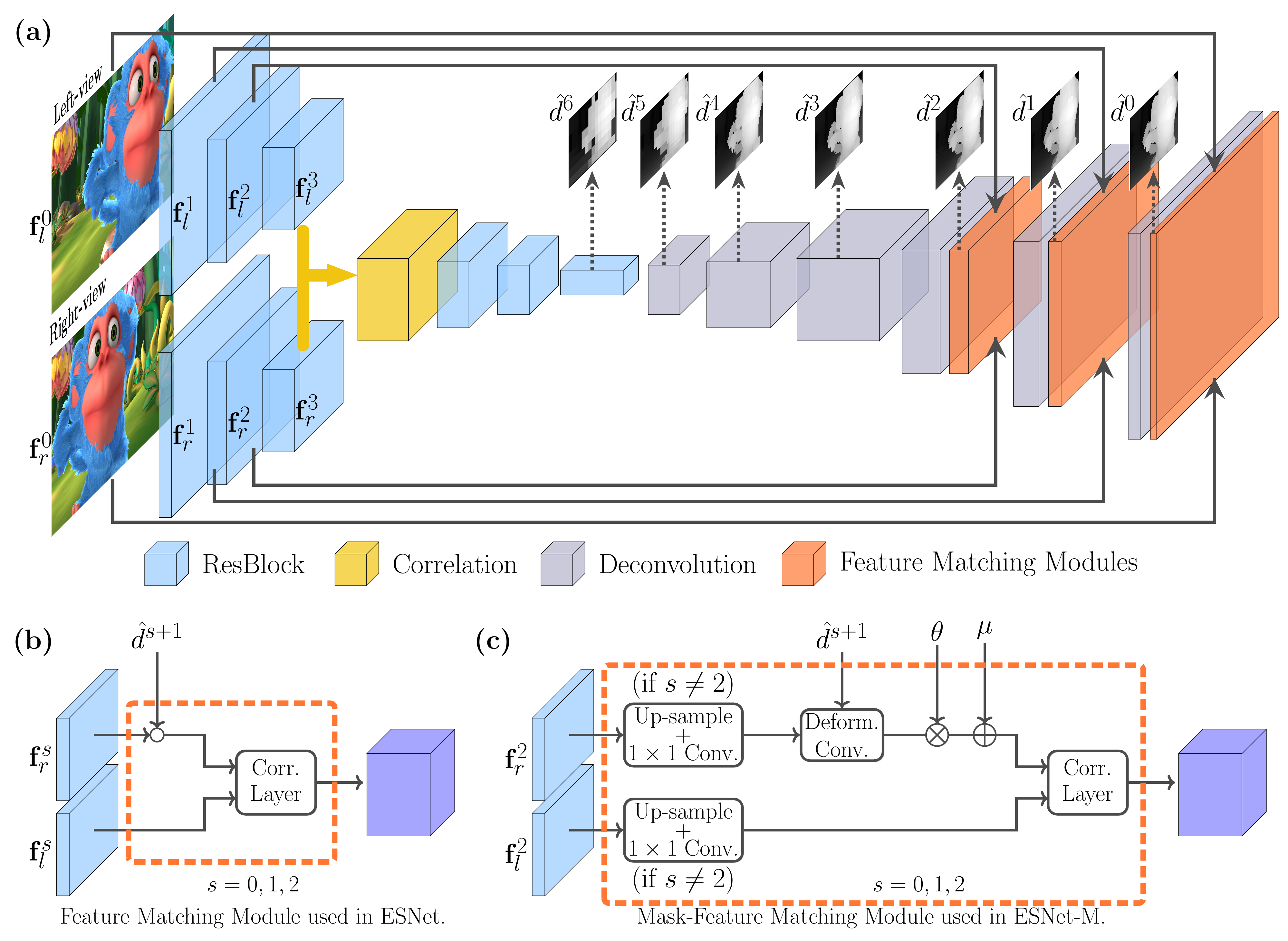}
\caption{Network architecture and feature matching modules. (a) Network structure of \RBDispNetMC, with the details of its feature matching module shown in (b). The network outputs disparity at multiple scales: \mbox{$\{\hat{d}^s:s = 0\dots, 6\}$}. UNet-like skip-connections~\cite{Unet} are not drawn for clarity; The architecture of \RBDispNetMOC is similar to (a), but uses the feature matching module shown in (c).}
\label{fig:Network Structure}%
\vspace{-0.4cm}
\end{figure*}

\subsection{KEY COMPONENTS OF EFFICIENT STEREO MATCHING}
\label{subsec:Key components}
Here we describe the components we proposed for designing and training an efficient stereo matching network, including multi-scale cost volume with warping, occlusion modeling, unsupervised pre-training, and dataset scheduling. 

\subsubsection{\textbf{Multi-scale Cost Volume with Warping}} 
\label{subsec:Multi-scale}
\mbox{\RBDispNetC} uses a shared feature extractor similar to the one shown in Fig.~\ref{fig:Network Structure}(a) to extract multi-scale feature maps \mbox{$\{(\mathbf{f}_{l}^{s}, \mathbf{f}_{r}^{s}):s=0,1,2,3\}$} from the left-view ($l$) and the right-view ($r$) images, where the superscript $s$ indicates the scale level with spatial downsampling rate of $2^{s}$. The cost volume $c$ at pixel location $(x_1, y_1)$ is then computed from the feature maps at scale level $s=3$ using Eq.~(\ref{eq:cost_computation}) as: 
\begin{equation}
\label{eq:cost_computation}
%  c(\mathbf{x}_{1}, \mathbf{x}_{2})=\frac{1}{N}\sum\langle\mathbf{f}_{1}(\mathbf{x}_{1}), \mathbf{f}_{2}(\mathbf{x}_{2})\rangle
 c(x_1, y_1, d)=\frac{1}{N_c}\langle\mathbf{f}_{l}(x_1, y_1), \mathbf{f}_{r}(x_1-d, y_1)\rangle,
\end{equation}
where $d$ is an integer within the maximum disparity search range, i.e., $d\in[0,D_{\text{max}})$, $\langle\dots\rangle$ is the dot product operation, and $N_c$ is the size of the feature channel. For feature maps that are at $s=3$, the size of computed cost volume is $\frac{W}{8} \times \frac{H}{8}\times D_{\text{max}}$, where $H$ and $W$ are the height and width of the input image, respectively. We choose $D_{\text{max}}=40$, which corresponds to a search range of 320 pixels at the original image resolution. \comment{Computing the cost volume using down-sampled feature maps enables matching pixel with large disparity, but at the cost of reduced spatial resolution and larger error.} To solve the problem of matching large pixel motions that may beyond $D_{\text{max}}$ and associated huge computational complexity, PWC-Net~\cite{PWCNet} proposes to warp the feature map from the source frame to the target frame before the correlation operation. Inspired by this idea, our proposed \RBDispNetMC efficiently calculates high resolution cost volumes through warping in a multi-scale fashion. As shown in Fig.~\ref{fig:Network Structure}(a), the network estimates disparity $d$ at multiple scales: \mbox{$\{\hat{d}^s:s = 0\dots, 6\}$} and computes multi-scale cost volume with $s=0,1,2$ using the Feature Matching Module (FMM). At each scale level $s$ and location $(x,y)$, FMM warps the extracted right-view feature map $\mathbf{f}_{r}^{s}$ using the bilinear-upsampled coarse disparity $\hat{d}^{(s+1)}$ from scale level $s+1$ as:
\begin{equation}
\label{eq:warping_disp}
\Tilde{\mathbf{f}}^{s}_r(x,y)=\mathbf{f}_{r}^{s}\Big(x+\text{up}_{2}\big(\hat{d}^{s+1}\big)\big(x,y\big), y\Big),    
\end{equation}
where $\Tilde{\mathbf{f}}^{s}_r$ is the warped right-view feature map and $\text{up}_2$ is the $\times2$ bilinear interpolation operation. The warped feature map $\Tilde{\mathbf{f}}^{s}_r$ is then correlated with the left-view feature using Eq.~\ref{eq:cost_computation}. The resulting cost volume from the correlation is then concatenated into the network and enables more accurate disparity estimation at subsequent scale levels. By our design, large pixel motions between the two views are compensated by warping, and the correlation operation can serve as a fine-tuning step given the coarse matching from warping. Hence, the search range needed in Eq.~\ref{eq:cost_computation} is significantly reduced (we use $d\in[-2,2]$ for all experiments and using a larger search range doesn't improve the accuracy). In addition, such design of multi-scale cost volume enables efficient and accurate matching at multiple-scales, where low resolution feature maps focus on global and large motion (as the receptive field is larger), and high resolution feature maps focus on local and fine-detailed motion (as the receptive field is smaller).

\begin{figure*}[tb!]
\vspace{7pt}
%\captionsetup{width=0.99\textwidth}
\begin{minipage}[b]{1.0\linewidth}
  \centering
  \setlength\tabcolsep{1.5pt}
  {\renewcommand{\arraystretch}{0.5}% for the vertical padding
    \begin{tabular}{ccccc} 
    % Image & PWC-Net-C5 (s=6) & PWC-Net-C5 (s=4) & PWC-Net-C5 (s=2) \\
    \multirow{2}{*}[1.2cm]{{\begin{tikzpicture}
    \node[inner sep=0pt, label=above:{Image}] at (0,0) {\includegraphics[width =0.24\linewidth, height =0.1183\linewidth, trim = {2.1cm 2.5cm 2cm 3.12cm}, clip]{{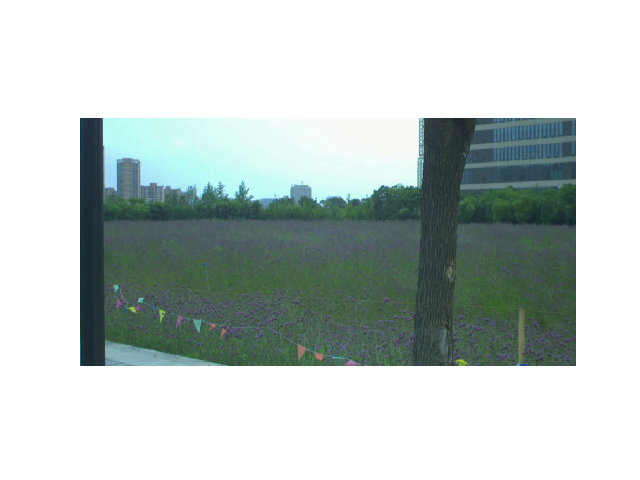}}};
    \end{tikzpicture}}} & 
    {\begin{tikzpicture}
    \node[inner sep=0pt, label=above:{PWC-Net-C5 ($s=6$)}] at (0,0) {\includegraphics[width =0.24\linewidth, height =0.1183\linewidth, trim = {2.1cm 2.5cm 2cm 3.12cm}, clip]{{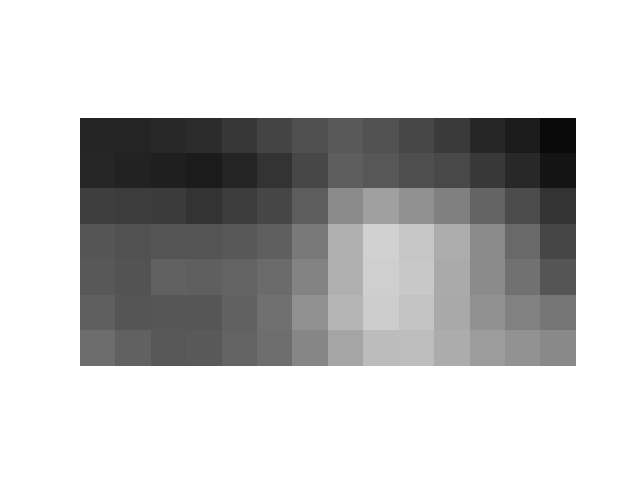}}};
    \draw[red,thick] (-0.1,-0.9) rectangle (0.8,0);
    \end{tikzpicture}} &
    {\begin{tikzpicture}
    \node[inner sep=0pt, label=above:{PWC-Net-C5 ($s=4$)}] at (0,0) {\includegraphics[width =0.24\linewidth, height =0.1183\linewidth, trim = {2.1cm 2.5cm 2cm 3.12cm}, clip]{{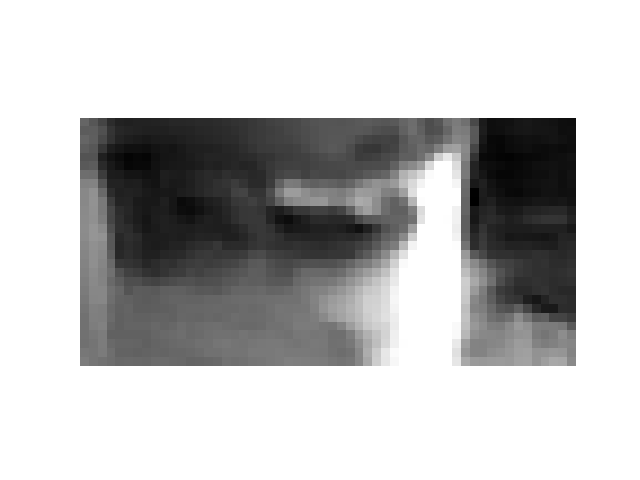}}};
    \draw[red,thick] (-0.1,-0.9) rectangle (0.8,0);
    \end{tikzpicture}}&
    {\begin{tikzpicture}
    \node[inner sep=0pt, label=above:{PWC-Net-C5 ($s=2$)}] at (0,0) {{\includegraphics[width =0.24\linewidth, height =0.1183\linewidth, trim = {2.1cm 2.5cm 2cm 3.12cm}, clip]{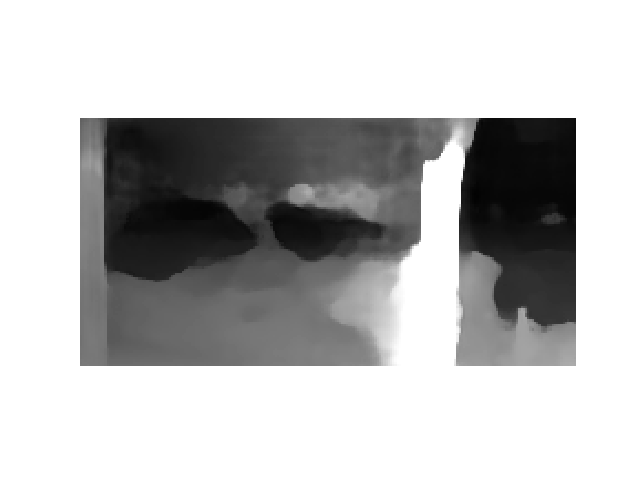}}};
    \draw[red,thick] (-0.1,-0.9) rectangle (0.8,0);
    \end{tikzpicture}} \\ &
    {\begin{tikzpicture}
    \node[inner sep=0pt, label=above:{PWC-Net-C3 ($s=4$)}] at (0,0) {{\includegraphics[width =0.24\linewidth, height =0.1183\linewidth, trim = {2.1cm 2.5cm 2cm 3.12cm}, clip]{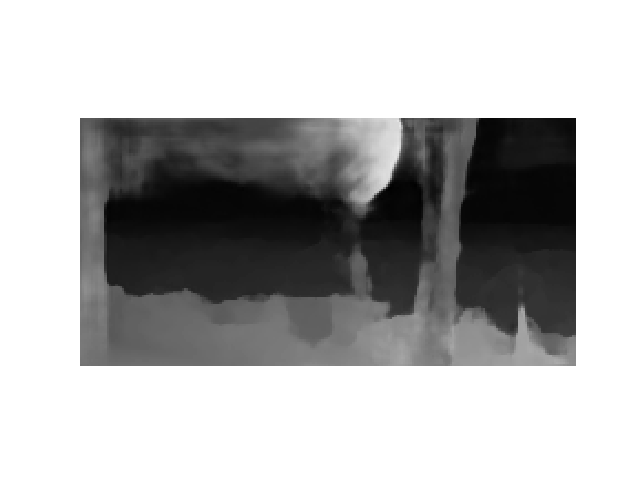}}};
    \draw[red,thick] (-0.1,-0.9) rectangle (0.8,0);
    \end{tikzpicture}} &
    {\begin{tikzpicture}
    \node[inner sep=0pt, label=above:{PWC-Net-C3 ($s=3$)}] at (0,0) {{\includegraphics[width =0.24\linewidth, height =0.1183\linewidth, trim = {2.1cm 2.5cm 2cm 3.12cm}, clip]{images/viz_high_res_corr/high_res_corr/112_224_3_corr_vmax_180.png}}};
    \draw[red,thick] (-0.1,-0.9) rectangle (0.8,0);
    \end{tikzpicture}} &
    {\begin{tikzpicture}
    \node[inner sep=0pt, label=above:{PWC-Net-C3 ($s=2$)}] (image) at (0,0) {{\includegraphics[width =0.24\linewidth, height =0.1183\linewidth, trim = {2.1cm 2.5cm 2cm 3.12cm}, clip]{images/viz_high_res_corr/high_res_corr/112_224_3_corr_vmax_180.png}}};
    \draw[red,thick] (-0.1,-0.9) rectangle (0.8,0);
    %\node[above=0 of image] {Hello world}
    \end{tikzpicture}}
    
    \end{tabular} 
    } 
\end{minipage}
\vspace{-2 pt}
\caption{Comparison of estimated disparity at different scales, using PWC-Net adapted for stereo matching computing cost volume on scale $s=6,5,4,3,2$ (PWC-Net-C5, top row) and on scale $s=4,3,2$ (PWC-Net-C3, bottom row). PWC-Net-C5 has large disparity error in the highlighted region in the low resolution output and the error is persistent in the high resolution output.}
\vspace{-5pt}
\label{fig:high_res_corr}
\end{figure*}

The proposed \RBDispNetMC uses a warping and correlation operation similar to PWC-Net, but with an important difference: our proposed \RBDispNetMC computes multi-scale cost volumes at scale levels \mbox{$s=2,1,0$} using FMM, instead of all scales as in PWC-Net. This is because coarsely matching pixels at extremely low resolution can introduce errors that are difficult to correct in later high resolution estimations. \fig{\ref{fig:high_res_corr}} illustrates this point by showing the estimated disparities at multiple scales from a normal PWC-Net adapted for stereo matching computing cost-volume at scale $s=6,5,4,3,2$ and a modified PWC-Net only computing cost-volume at high resolution scale $s=4,3,2$. The estimated disparity of PWC-Net-C5 at $s=6$ shows blurred edges around the tree and this error propagates to higher resolution estimates. \comment{Computing cost volume at high resolution ensures accurate disparity estimation and the warping operation makes the cost volume computation efficient even at high resolution.} By only calculating cost-volumes at high resolutions, we are able to improve the accuracy and reduce the runtime at the same time. The second row of table~\ref{table:Comparison:multi-scale vs single scale and MFN} shows that our proposed \RBDispNetMC achieves lower end point error (EPE) on both of the Scene Flow and the DrivingStereo dataset than the baseline method.

\begin{table*}[tb!]
\vspace{7 pt}
\caption{Ablation studies for the effect of multi-scale cost volume, occlusion modeling, and unsupervised pre-training on different models evaluated on the Scene Flow and DrivingStereo datasets. The End Point Error (EPE) is used for evaluation.}
\label{table:Comparison:multi-scale vs single scale and MFN}
\begin{center}
\begin{tabular}{|c|c|c|c|c|c|}
\hline  { Model } & {Multi-scale cost volume}& {Occlusion Modeling}& {Unsupervised pre-training}& {Scene Flow } &  { DrivingStereo } \\

\hline Baseline~\cite{FADNet} & - & - &- & 1.21 & \comment{0.62}0.68  \\

\hline \multirow{2}{*}{\RBDispNetMC} &\checkmark&- &- & 1.18 & \comment{0.53}0.61  \\
\cline{2-6}  &\checkmark&- & \checkmark & 1.03 & \comment{0.51}0.59  \\
\hline \multirow{2}{*}{\RBDispNetMOC} &\checkmark&\checkmark &-& ${1.03}$ &\comment{0.52}0.59  \\
\cline{2-6}  &\checkmark&\checkmark & \checkmark& $\mathbf{0.95}$ & \comment{0.49}\textbf{0.55}  \\

\hline
\end{tabular}
\end{center}
\vspace{-0.2cm}
\end{table*}

\subsubsection{\textbf{Occulusion Modeling}}
\label{subsec:Occlusion}\begin{figure}[tb!]

\centering
\vspace{2 pt}
\begin{tikzpicture}[scale=1]

\node[] (im_r) at (0,0) {\includegraphics[width=3cm,trim = {2.5cm 2.5cm 7cm 2.7cm}, clip]{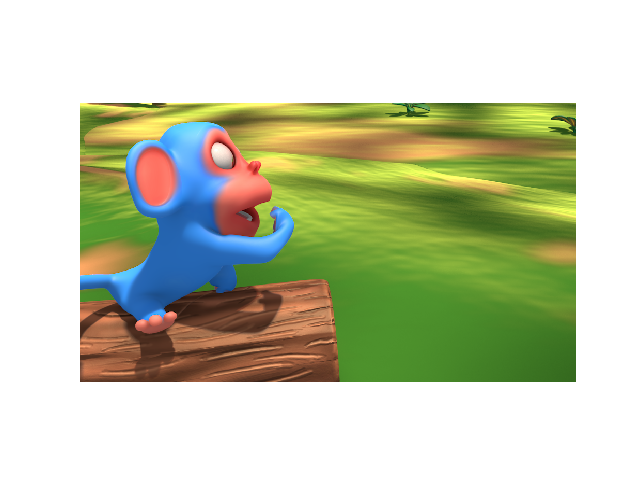}};
\node[] (im_r_warped) at (5,0) {\includegraphics[width=3cm,trim = {2.5cm 2.5cm 7cm 2.7cm}, clip]{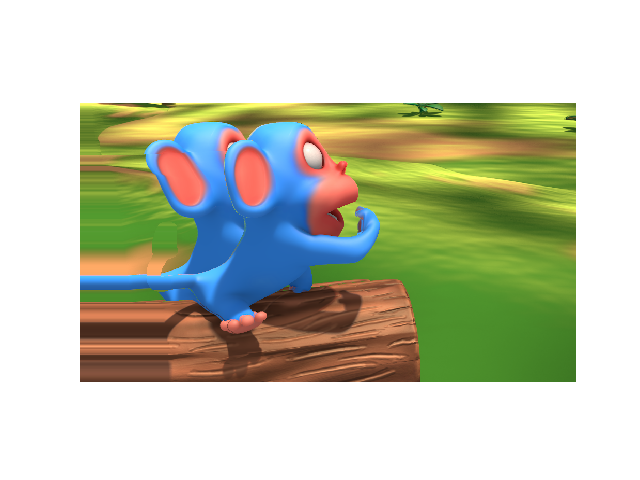}};
\draw[-stealth, line width = 3pt] (im_r) -- (im_r_warped) node [midway, above] {Warping};

\end{tikzpicture}
\caption{Illustration of ghosting effect in occluded regions, where the warped right-view image contains duplicated monkeys.}
\label{fig:ghosting effect}
\vspace{-0.5cm}
\end{figure}
Warping operation described in Section~\ref{subsec:Multi-scale} is able to compensate for the pixel shift due to disparity and make the warped right-view image similar to the left-view image. However, this is only true for non-occluded regions. When occlusion occurs, the warped image may show duplicated objects, also known as `ghosting effect'. This is illustrated in Fig.~\ref{fig:ghosting effect}, where the warped right-view image contains duplicated monkeys. Duplicated objects in the warped image due to occlusion can negatively affect the stereo matching performance since it would be ambiguous about which of the duplicated objects in the warped right-view image should be matched with the object in the left-view. 

To alleviate such issue, we propose \RBDispNetMOC, a variant of \RBDispNetMC, that has better occlusion modeling as illustrated in Fig.~\ref{fig:Network Structure}. It shares the same overall network structure with \RBDispNetMC and has an occlusion-aware feature matching module inspired by  MaskFlowNet~\cite{MFN} , which we call Mask-Feature Matching Module (Mask-FMM). 

At each scale level $s$ $(s=0,1,2)$, Mask-FMM takes in feature map $\mathbf{f}_{l}^{2}$ and $\mathbf{f}_{r}^{2}$. For scale $s\neq 2$, bilinear-upsampling and $1 \times 1$ convolution are used to transform $\mathbf{f}_{l}^{2}$ and $\mathbf{f}_{r}^{2}$  to proper spatial and channel dimension.  Next, it warps the right-view feature map by up-sampled coarse disparity estimate $\hat{d}^{s+1}$ using deformable convolution~\cite{DeformableConv}. The warped feature is then element-wise multiplied ($\otimes$) by a soft learnable occlusion mask~$\theta$ of shape $(H, W)$ by broadcasting and added element-wise ($\oplus$) by a trade-off term~$\mu$ of shape \mbox{$(C, H, W)$}. The occlusion mask $\theta$ and the trade-off term $\mu$ here are both derived from convolution outputs at scale $s+1$. By filtering the feature map with occlusion mask $\theta$ and supplying the masked feature region with $\mu$, it improves the quality of the cost volume and alleviates the ghosting effect due to occlusion. Importantly, the rough occlusion mask $\theta$ is learned without supervision and the correct gradient flow for $\theta$ is completely due to the lower loss when occluded regions are masked and hence resulting in better cost volumes. 

Note that the input to the proposed Mask-FMM module at $s=0,1$ are both \mbox{($\mathbf{f}_{l}^{2}$,$\mathbf{f}_{r}^{2})$}. We found that using such high-semantic level feature map (with upsample and $1\times 1$ \text{Conv.}) gives better occlusion estimation compared to using \mbox{($\mathbf{f}_{l}^{1}$,$\mathbf{f}_{r}^{1})$} at $s=1$ and \mbox{($\mathbf{f}_{l}^{0}$,$\mathbf{f}_{r}^{0})$} at $s=0$. Our approach leads to high resolution, semantic-rich feature maps, which shares the same spirit with feature pyramid network~\cite{FPN}. Such high-semantic level feature maps are critical to ensure a good occlusion estimation. Since in the opposite extreme, a very poor quality feature maps would force the network to use the trade-off feature solely in correlation, by learning a completely occluded (black) occlusion map.

Table~\ref{table:Comparison:multi-scale vs single scale and MFN} shows that \RBDispNetMOC achieves higher accuracies than \RBDispNetMC on both of the Scene Flow dataset and the DrivingStereo dataset (compare the third and the fifth row). It is observed that the improvement of \RBDispNetMOC on Scene Flow is more significant, possibly because occlusion is more prevalent in the Scene Flow dataset.

\subsubsection{\textbf{Unsupervised Pre-training}}
\label{subsec:Unsupervised_pretrain}
Unsupervised pre-training has been shown to be helpful in improving network performance on the tasks of image classification~\cite{Unsupervised_pretrain}\cite{MOCO}, flow estimation \cite{UnFlow} and object detection~\cite{MOCO} through learning useful feature embedding that serves as better initialization for supervised training. But to the best of our knowledge, no prior work has applied unsupervised pre-training to the problem of stereo matching. Since unsupervised training only introduces computation cost during training time and does not impact the inference speed, we explore whether it will help to improve the performance for efficient stereo matching.

We use photometric reprojection loss similar to MonoDepth~\cite{MonoDepth} and MonoDepth2~\cite{MonoDepth2} to pre-train the stereo matching network in an unsupervised manner. Assuming the training loss $L^s$ at scale level $s$, the total loss $L$ is then given by $ L = \sum_{s=0}^{3} L^s$. Note that we only use 4 scales to calculate the loss $L$, as is done in MonoDepth2~\cite{MonoDepth2}. Each $L^s$ consists of a photo-reprojection error part and a disparity smoothness part, i.e., $L^s = \lambda_1 L_{pe}^s + \lambda_2 L_{sm}^s$ where $\lambda_1$ and $\lambda_2$ are loss weights. The photo-reprojection error $L_{pe}^s$ is given by:
\begin{equation}
    L^s_{pe} =\frac{1}{N}\sum\alpha \frac{1-SSIM(\Tilde{I}^s_r, I_l)}{2}+(1-\alpha)\|\Tilde{I}^s_r-I_l\|_{1},
\end{equation}
where $\alpha=0.85$, $SSIM$ is the structural similarity index measure~\cite{SSIM}. $I_l$ is the left-view image, $\Tilde{I}^s_r$ is the warped right-view image using $\hat{d}^s$, and $N$ is the number of pixels. The disparity smoothness term $L_{sm}^s$ is given by:
\begin{equation}
L_{sm}^s=\frac{1}{N} \sum\left|\partial_{x} \hat{d}^s \right| e^{-\left|\partial_{x} I^s_l\right|}+\left|\partial_{y} \hat{d}^s\right| e^{-\left|\partial_{y} I^s_l\right|},
\end{equation}
where $I^s_l$ is the down-sampled version of $I_l$ having the same size as $d^s$. In all experiments, we used
$\lambda_1 = 5$, $\lambda_2=0.1$ and pre-trained the networks for 30 epochs. Table~\ref{table:Comparison:multi-scale vs single scale and MFN} shows that unsupervised pre-training on the same dataset before supervised training consistently improves the disparity estimation accuracy for both \RBDispNetMC and \RBDispNetMOC. It should be noted that the improvement from proposed unsupervised pre-training is not due to the longer training of the model: the training of the model without unsupervised pre-training has already converged and additional training iterations won't improve the performance. Finally, we note that instead of unsupervised pre-training, it is also possible to directly train the network using a mixed loss that contains both supervised loss (\ref{subsec:Loss}) and our unsupervised loss. However, we found that it leads to less improvement in accuracy compared to our proposed unsupervised pre-training.

\subsubsection{\textbf{Dataset Scheduling}}
\label{subsec:Dataset Scheduling}
Training deep stereo matching networks requires a large amount of training data. A common practice is to first train on the synthetic Scene Flow dataset, and then finetune on the specific dataset of interest, e.g., KITTI. This is mostly because collecting a large realistic dataset with accurate depth ground truth is challenging and time-consuming. However, with the recently introduced large scale dataset DrivingStereo~\cite{DrivingStereo}, a natural question to ask is whether additional training using DrivingStereo dataset is helpful to improve the network's performance on the KITTI benchmark and what would be the best dataset schedule for training. \comment{Similar question has been answered in the problem of optical flow estimation~\cite{mayer2018DatasetSchedule}\cite{PWCNet}, where it shows that training using specific order of the datasets yields best performance.}\comment{i.e., training in the order of Scene Flow and then DrivingStereo, or training in the order of DrivingStereo and then Scene Flow.} In this paper, we re-examine the standard training procedure and evaluate the models trained with different dataset schedules on the KITTI 2015 dataset. The result in Table~\ref{table:Comparison:Dataset schedule} shows that training sequentially on Scene Flow (SF), then on DrivingStereo (DS), and finally on KITTI (K) gives the lowest error rate for all models, indicating that the inclusion of DrivingStereo dataset for training is very helpful. It is also worth noting that training in the order of DS+SF+K gives worse performance compared to the order SF+DS+K. Since DrivingStereo is much more similar to the KITTI dataset, this may indicate that the scheduled dataset that is more similar to the target dataset should be used later in the training schedule. We expect our proposed training methods described in \textbf{\it{3)}} and \textbf{\it{4)}} can be helpful to improve the performance of other models as well.
\begin{table}[tb!]
\vspace{5pt}
\caption{Error rate comparison of models trained with different dataset schedules on KITTI 2015 validation dataset. The error rate is defined as the percentage of disparity estimation with EPE $\geq 3$ pixels or $\geq 5\%$.}
\label{table:Comparison:Dataset schedule}
\begin{center}
\begin{tabular}{|c|c|c|}
\hline
Model & Dataset Schedule & Error Rate \\ \hline

\multirow{4}{*}{$\text{\RBDispNetMC}$} & SF+K &2.80\%  \\ \cline{2-3} 
 & DS+K &2.14\%  \\ \cline{2-3} 
 & DS+SF+K & ${2.11}\%$  \\ \cline{2-3}
 & SF+DS+K & $\mathbf{2.02\%}$  \\ \hline
\multirow{4}{*}{$\text{\RBDispNetMOC}$} & SF+K & 2.38\% \\ \cline{2-3} 
 & DS+K & 2.36\% \\ \cline{2-3} 
 & DS+SF+K & 2.03\% \\ \cline{2-3} 
 & SF+DS+K & $\mathbf{1.85\%}$ \\ \hline

\end{tabular}
\end{center}
\vspace{-0.5cm}
\end{table}

\subsection{Loss Function}
\label{subsec:Loss}
Here we describe the loss function we used for training networks with supervision. Given the multi-scale disparity \mbox{$\{\hat{d}^s:s = 0\dots, 6\}$} estimated by the proposed networks, we trained the networks by minimizing multi-scale smooth $L_1$ loss similar to~\cite{SceneflowAndDispNetC}\cite{FADNet}. The loss $L^s$ at scale $s$ is given by:
\begin{equation}
\begin{aligned}
&L^s(\hat{d}^s, d^s) =\frac{1}{N} \sum l(|\hat{d}^s-d^s|) \\
&l(x) \triangleq \left\{\begin{array}{ll}
x^{2} / 2, & |x|<1 \\
|x|-0.5, &\text{otherwise.}

\end{array}\right.
\end{aligned}  
\end{equation}
where $d^s$ is the ground truth disparity down-sampled to scale $s$, and $N$ is the number of pixels. The total loss $L$ is calculated as a weighted sum of $L^s$:
\begin{equation}
    L = \sum_{s=0}^{6} \omega^s L^s,
\end{equation}
where $\{\omega^s:s=0\dots,6\}$ are scale-dependent weights.

\begin{figure*}[tb!]
\vspace{7pt}

\begin{minipage}[b]{1.0\linewidth}
  \centering
  \setlength\tabcolsep{1.5pt}
  {\renewcommand{\arraystretch}{0.5}% for the vertical padding
    \begin{tabular}{cccccc} 
    \small{Image} & \small{Ground Truth} & \small{\RBDispNetC} & \small{\RBDispNetMOC (ours)} &  \small{Estimated occlusion mask}\\
    
    {\includegraphics[width =0.19\linewidth, trim = {2.1cm 2.3cm 2cm 2.3cm}, clip]{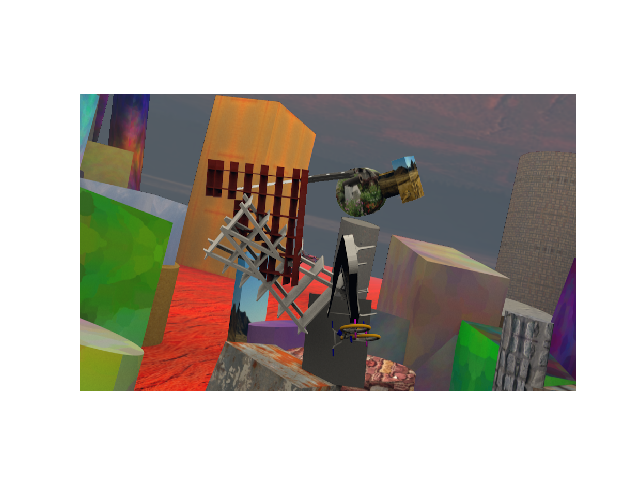}} &
    {\includegraphics[width =0.19\linewidth, height =0.1183\linewidth, trim = {2.1cm 2.55cm 2cm 2.55cm}, clip]{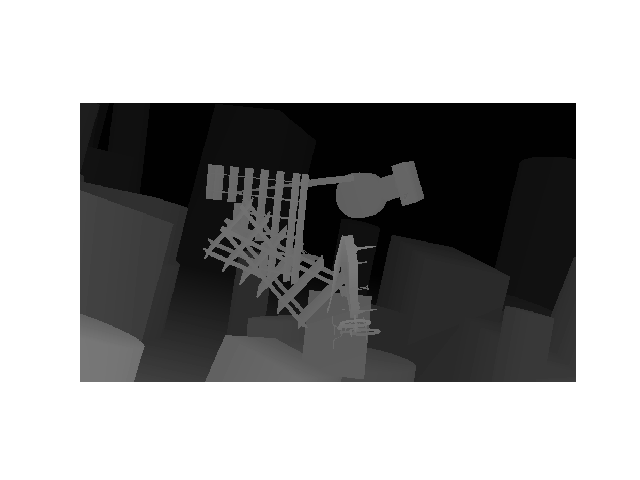}} &
    {\begin{tikzpicture}
    \node[inner sep=0pt] at (0,0) {\includegraphics[width =0.19\linewidth, height =0.1183\linewidth, trim = {2.1cm 2.5cm 2cm 3.12cm}, clip]{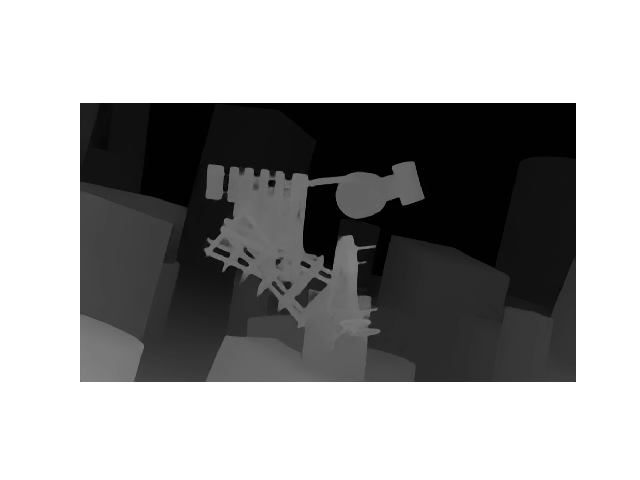}};
    \draw[red,thick] (-0.8,-0.4) rectangle (-0,0.6);
    \end{tikzpicture}}&
    {\begin{tikzpicture}
    \node[inner sep=0pt] at (0,0) {{\includegraphics[width =0.19\linewidth, height =0.1183\linewidth, trim = {2.1cm 2.5cm 2cm 3.12cm}, clip]{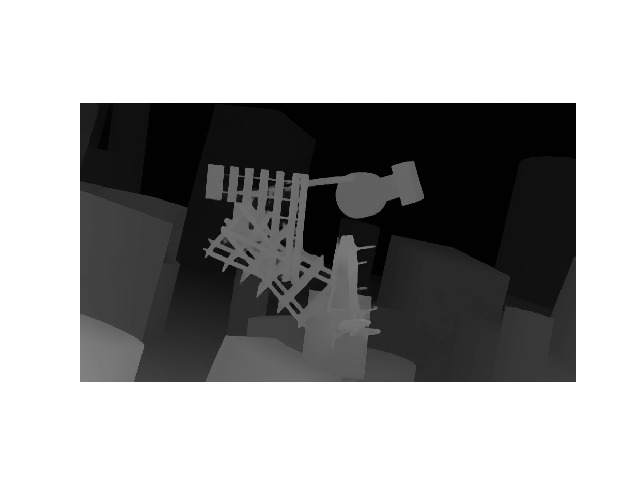}}};
    \draw[red,thick] (-0.8,-0.4) rectangle (-0,0.6);
    \end{tikzpicture}}&
    {\includegraphics[width =0.19\linewidth, trim = {2.1cm 2.3cm 2cm 2.3cm}, clip]{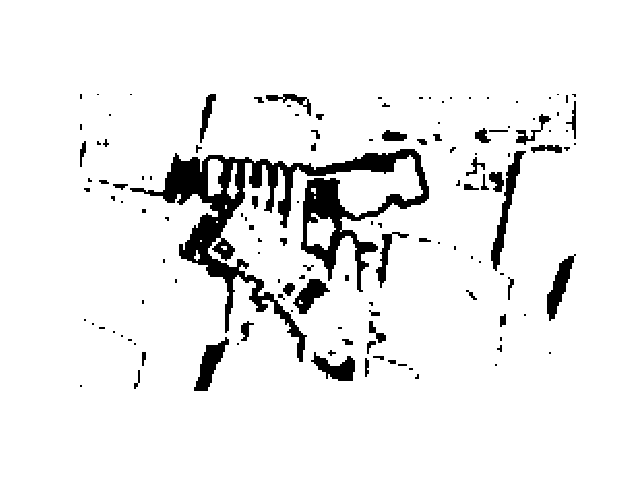}}    \\

    {\includegraphics[width =0.19\linewidth, trim = {2.1cm 2.3cm 2cm 2.3cm}, clip]{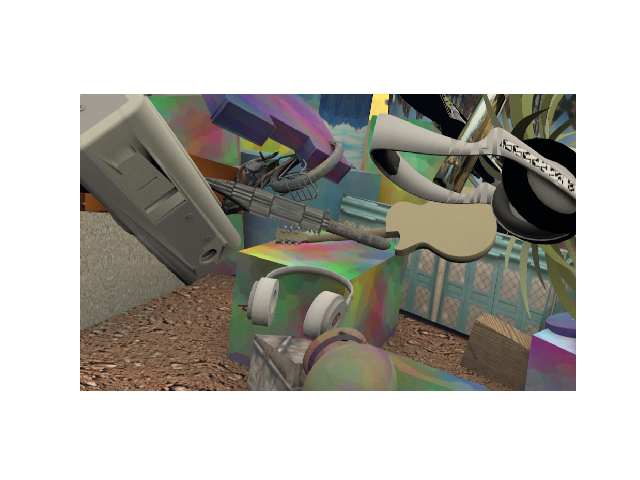}} &
    {\includegraphics[width =0.19\linewidth, height =0.1183\linewidth, trim = {2.1cm 2.55cm 2cm 2.55cm}, clip]{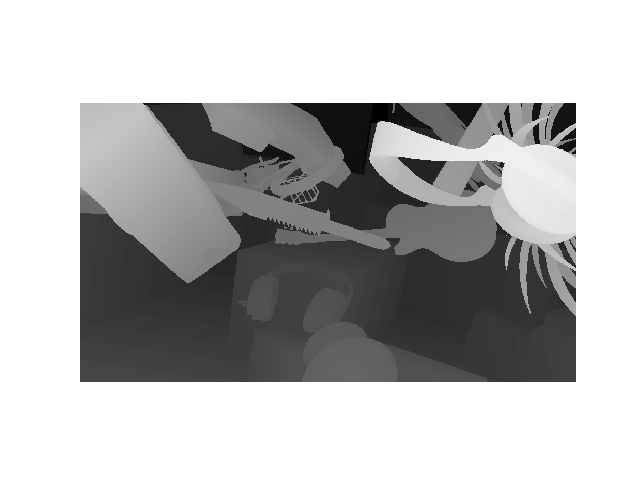}} &
    {\begin{tikzpicture}
    \node[inner sep=0pt] at (0,0) {\includegraphics[width =0.19\linewidth, height =0.1183\linewidth, trim = {2.1cm 2.5cm 2cm 3.12cm}, clip]{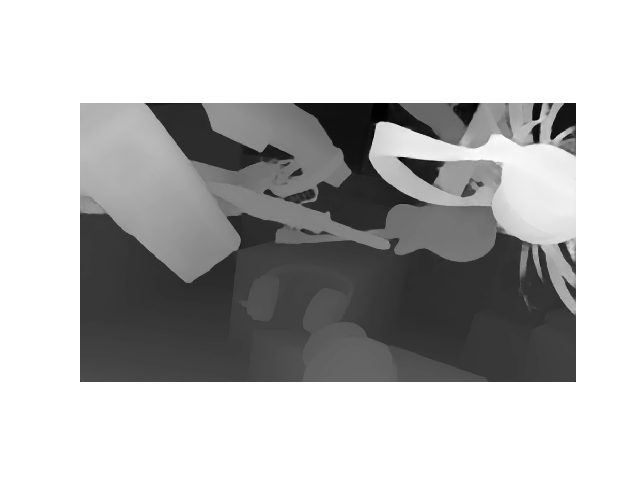}};
    \draw[red,thick] (-0.4,0.25) rectangle (0,0.55);
    \draw[red,thick] (1.3,-0.55) rectangle (1.65,0.1);
    \end{tikzpicture}}&
    {\begin{tikzpicture}
    \node[inner sep=0pt] at (0,0) {\includegraphics[width =0.19\linewidth, height =0.1183\linewidth, trim = {2.1cm 2.5cm 2cm 3.12cm}, clip]{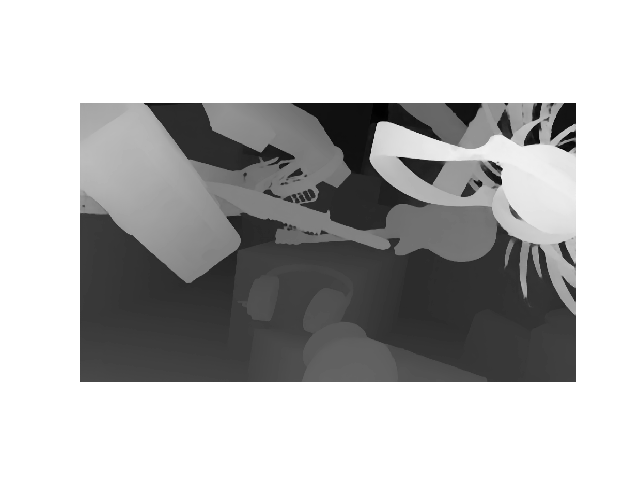}};
    \draw[red,thick] (-0.4,0.25) rectangle (0,0.55);
    \draw[red,thick] (1.3,-0.55) rectangle (1.65,0.1);
    \end{tikzpicture}}&
    {\includegraphics[width =0.19\linewidth, trim = {2.1cm 2.3cm 2cm 2.3cm}, clip]{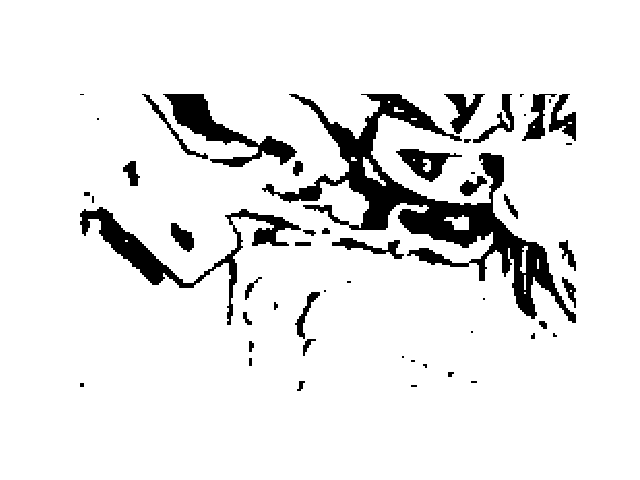}}

    \end{tabular}
    }
\end{minipage}
\vspace{0 pt}
\caption{Visual comparison of our proposed \RBDispNetMOC with \RBDispNetC on Scene Flow test set.}
\vspace{-5pt}
\label{fig:Ablation_Scene Flow}
\end{figure*}

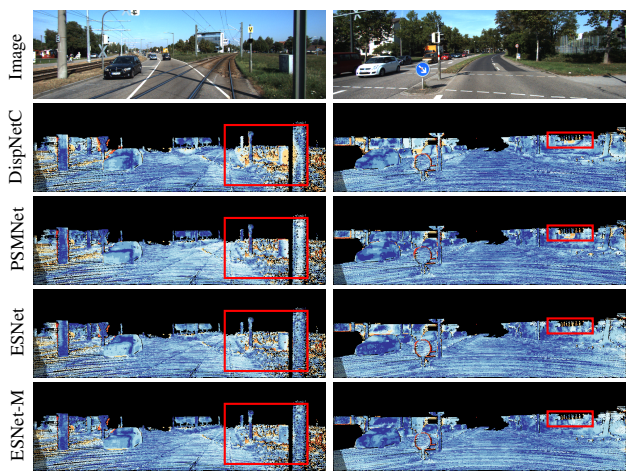
\begin{figure}[tb!]
%\captionsetup{width=0.99\textwidth}
\begin{minipage}[b]{1.0\linewidth}
  \centering
  \setlength\tabcolsep{1.5pt}
  {\renewcommand{\arraystretch}{0.5}% for the vertical padding   
    \begin{tabular}{ccc}
    \rotatebox[origin=l]{90}{\hspace{2mm}\scriptsize Image}&
    {\includegraphics[width =0.45\linewidth]{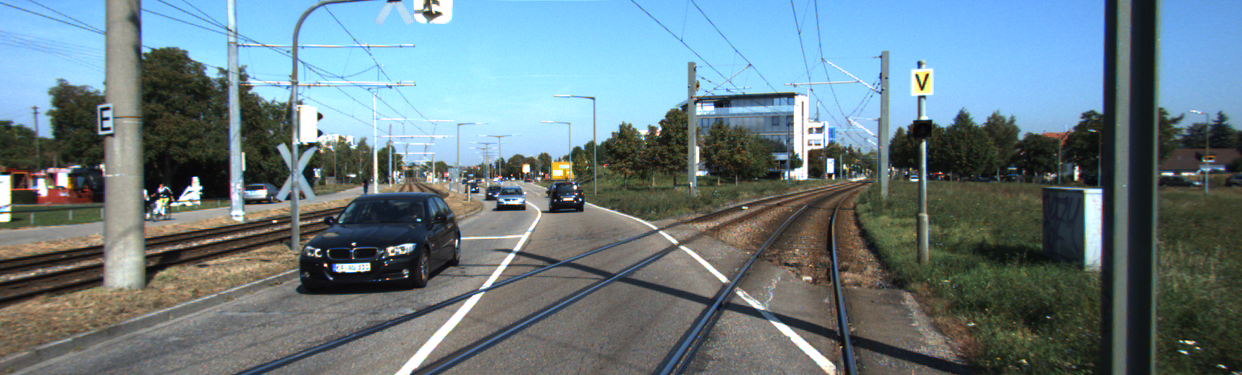}}&
    {\includegraphics[width =0.45\linewidth]{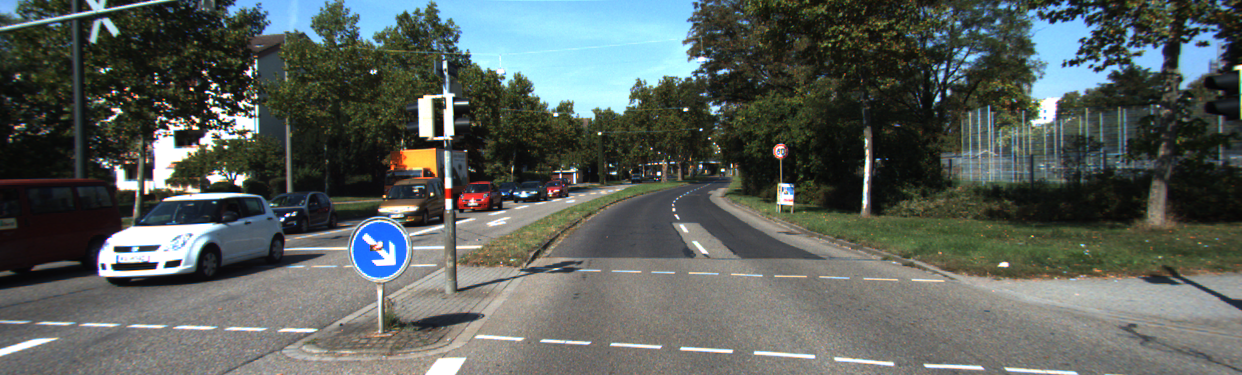}}\\

    \rotatebox[origin=l]{90}{\hspace{0.5mm}\scriptsize DispNetC}&
    {\begin{tikzpicture}
    \node[inner sep=0pt] at (0,0) {\includegraphics[width =0.45\linewidth]{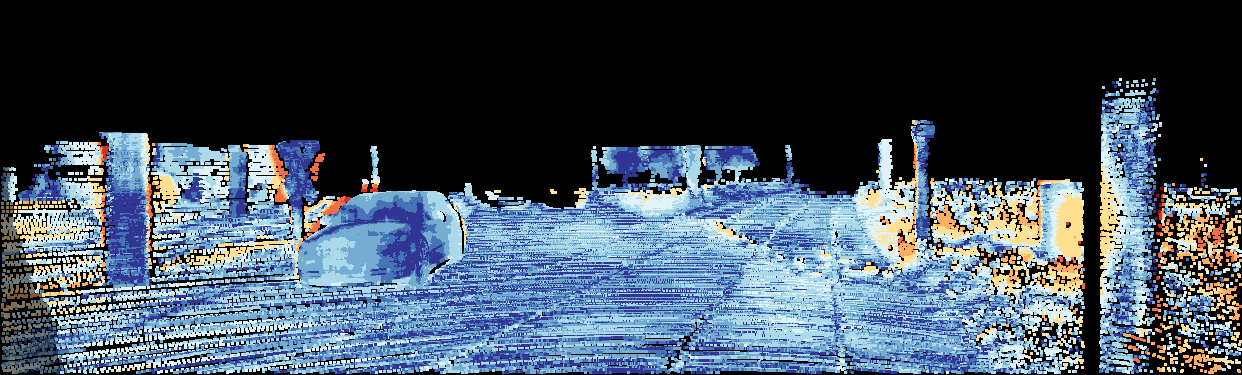}};  
    \draw[red,thick] (0.6,-0.5) rectangle (1.7,0.3);
    \end{tikzpicture}} &
    {\begin{tikzpicture}
    \node[inner sep=0pt] at (0,0) {\includegraphics[width =0.45\linewidth]{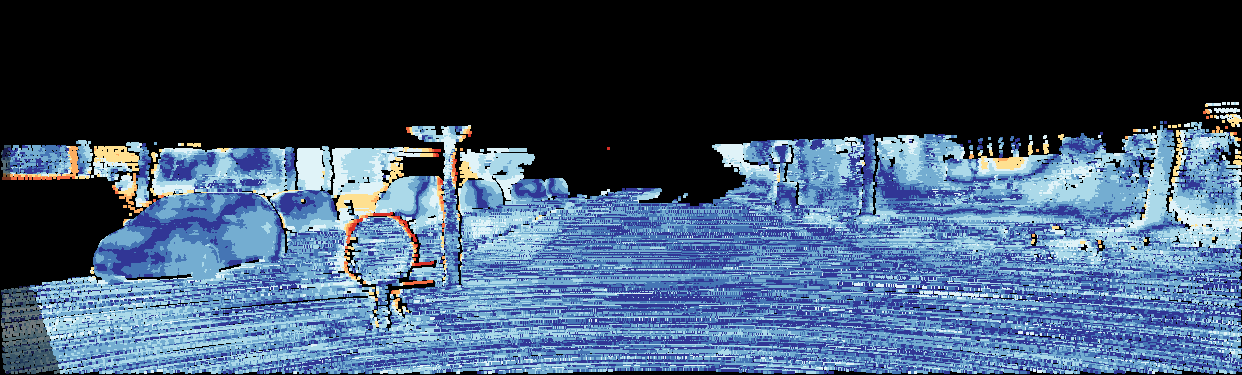}};
    \draw[red, thick] (0.9,0.0) rectangle (1.5,0.2);
    \end{tikzpicture}}\\
    \rotatebox[origin=l]{90}{\hspace{1mm}\scriptsize PSMNet}&
    {\begin{tikzpicture}
    \node[inner sep=0pt] at (0,0) {\includegraphics[width =0.45\linewidth]{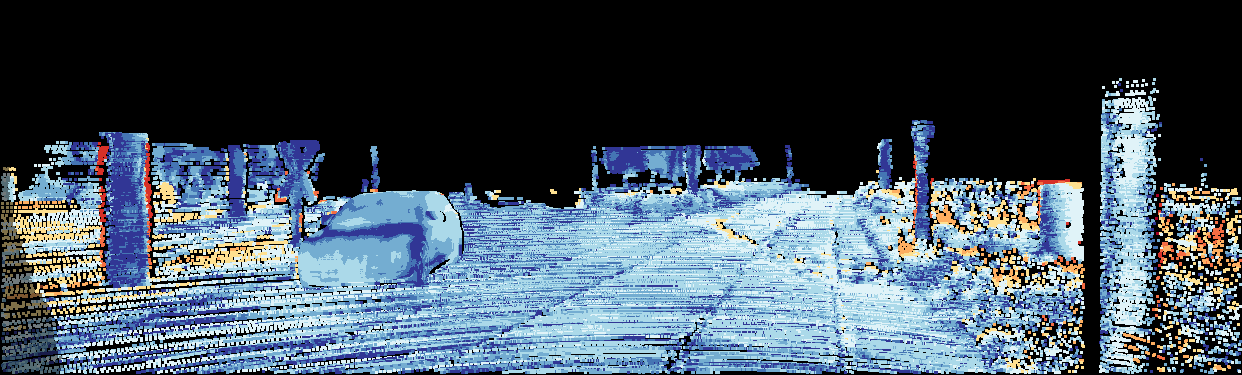}};  
    \draw[red,thick] (0.6,-0.5) rectangle (1.7,0.3);
    \end{tikzpicture}} &
    {\begin{tikzpicture}
    \node[inner sep=0pt] at (0,0) {\includegraphics[width =0.45\linewidth]{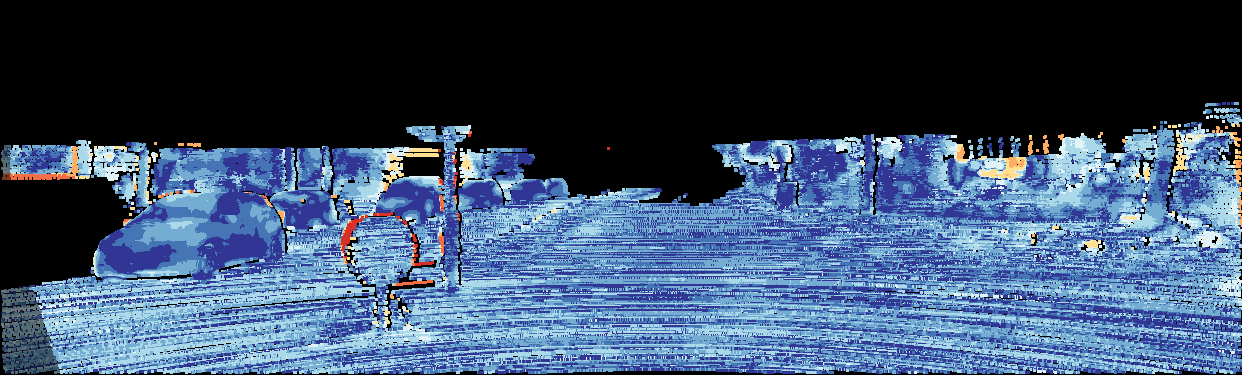}};
    \draw[red, thick] (0.9,0.0) rectangle (1.5,0.2);
    \end{tikzpicture}}\\
    \rotatebox[origin=l]{90}{\hspace{2mm}\scriptsize \RBDispNetMC}&
    {\begin{tikzpicture}
    \node[inner sep=0pt] at (0,0) {\includegraphics[width =0.45\linewidth]{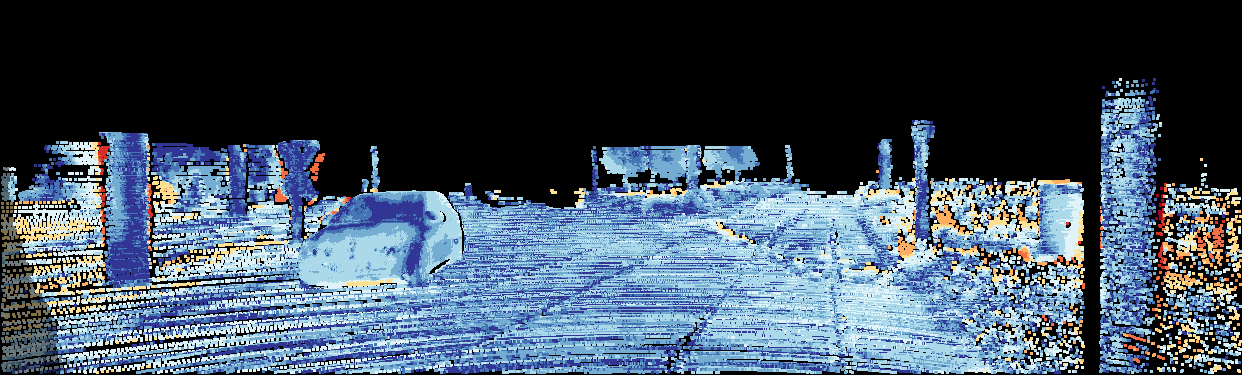}};  
    \draw[red,thick] (0.6,-0.5) rectangle (1.7,0.3);
    \end{tikzpicture}} &
    {\begin{tikzpicture}
    \node[inner sep=0pt] at (0,0) {\includegraphics[width =0.45\linewidth]{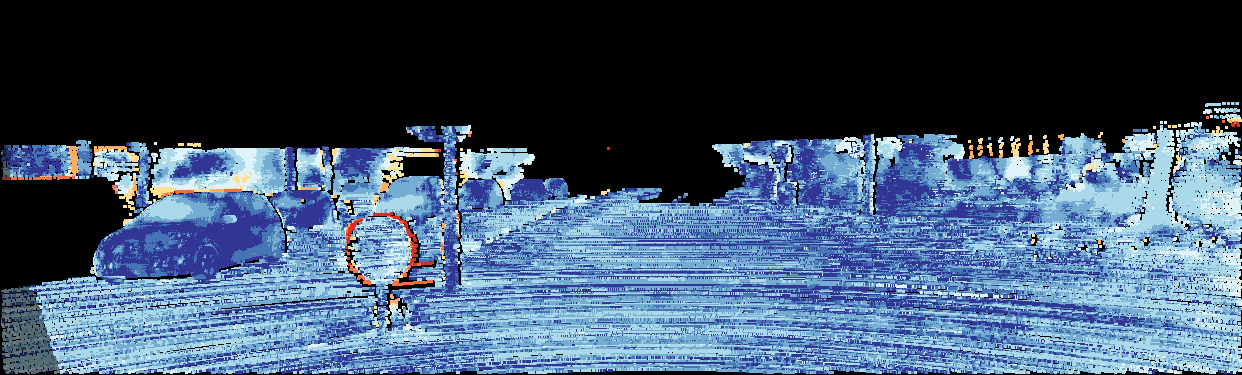}};
    \draw[red, thick] (0.9,0.0) rectangle (1.5,0.2);
    \end{tikzpicture}}\\    
    \rotatebox[origin=l]{90}{\hspace{0.5mm}\scriptsize \RBDispNetMOC}&
    {\begin{tikzpicture}
    \node[inner sep=0pt] at (0,0) {\includegraphics[width =0.45\linewidth]{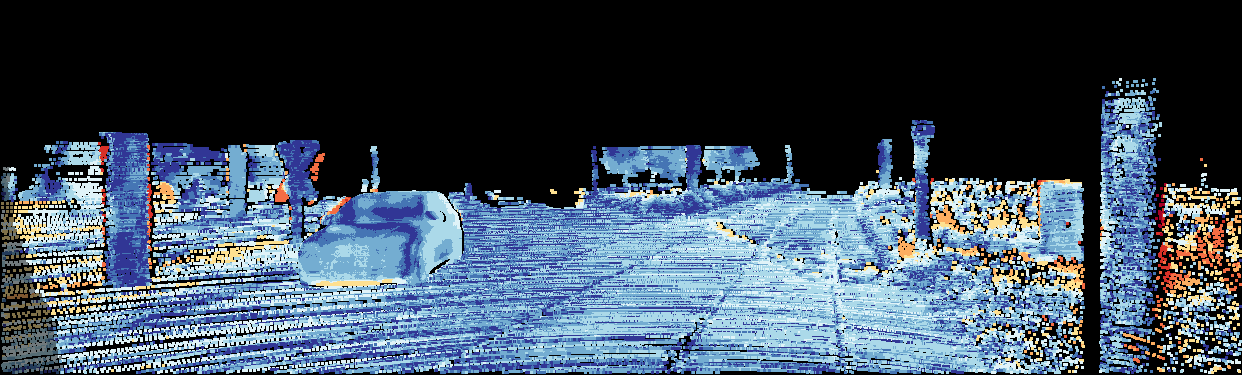}};  
    \draw[red,thick] (0.6,-0.5) rectangle (1.7,0.3);
    \end{tikzpicture}} &
    {\begin{tikzpicture}
    \node[inner sep=0pt] at (0,0) {\includegraphics[width =0.45\linewidth]{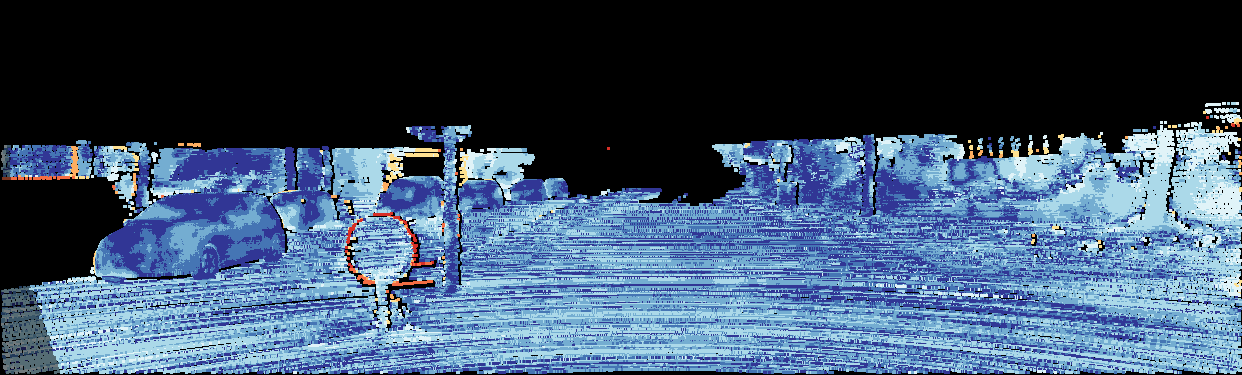}};
    \draw[red, thick] (0.9,0.0) rectangle (1.5,0.2);
    \end{tikzpicture}}\\

    \end{tabular}
    }

\end{minipage}
\caption{Visualization of disparity prediction error on the KITTI 2015 test set. Red and yellow denote large errors. Our proposed \RBDispNetMOC achieves similar accuracy as PSMNet while being $11 \times$ faster.}
\vspace{-5pt}
\label{fig:Comparison_KITTI}
\end{figure}

\section{EXPERIMENTS}
Here we compare the experimental results of our proposed networks with previously published stereo matching methods. We also report and compare the runtime of our networks with other networks. 

We use three publicly datasets, Scene Flow~\cite{SceneflowAndDispNetC}, DrivingStereo~\cite{DrivingStereo}, and KITTI 2015~\cite{KITTI2015}, to train and evaluate the performance of our proposed models. The Scene Flow dataset~\cite{SceneflowAndDispNetC} is a large synthetic dataset containing 35454 stereo image pairs for training and 4370 image pairs for testing. The DrivingStereo dataset~\cite{DrivingStereo} is a large realistic dataset containing 174437 training image pairs and 7751 testing image pairs for road driving scenarios with high density disparity labels obtained from model-guided filtering of multi-frame LiDAR points. The KITTI 2015 stereo dataset~\cite{KITTI2015} contains 200 training and 200 testing image pairs with ground truth disparity obtained from LiDAR and fitted 3D CAD models. We use 180 image pairs from the KITTI 2015 training set for fine-tuning and the remaining 20 pairs from the training set for validation.

\subsection{Implementation}
We implement our proposed networks using Pytorch~\cite{Pytorch}. The input images are normalized by ImageNet~\cite{ImageNet} mean ([0.485, 0.456, 0.406]) and standard deviation ([0.229, 0.224, 0.225]). The images during training are randomly cropped to size $384 \times 768$ for Scene Flow dataset, \mbox{$256 \times 768$} for DrivingStereo Dataset, and $256 \times 512$ for KITTI 2015 dataset. We train models using Adam optimizer~\cite{Adam} with $\beta_1=0.9$, $\beta_2=0.999$, batch size 16, on four Nvidia RTX 2080Ti GPUs. A multi-round training scheme following~\cite{FADNet} is used. Each round of training uses a different set of weights $\{\omega^s\}$ and the exact weights $\{\omega^s\}$ we used can be found in~\cite{FADNet}. For training on the Scene Flow dataset, a four-round training of 20, 20, 20, 30 epochs each is used. The learning rate is set to $10^{-4}$ at the beginning of each round and decays by half every 10 epochs. For training on the DrivingStereo dataset, we use a constant learning rate of $10^{-4}$ and trains the models by four rounds, with 7, 7, 7, 10 epochs each. For fine-tuning on the KITTI dataset, we apply a three-round training, with 1200 epochs each. The initial learning rate is set to $10^{-5}$ at the beginning of each round and is decayed by 10 at 600 epochs. For runtime analysis, we evaluate the inference time of exisiting models and our proposed models using a Nvidia RTX 2080Ti GPU using stereo images of size $576 \times 960$. 

\subsection{Results and Analysis}
For benchmark evaluations, we train our proposed \mbox{\RBDispNetMC} and \RBDispNetMOC with dataset schedule $\text{DS}^{\text{*}}$+DS+SF for evaluating on the Scene Flow dataset and trained with dataset schedule $\text{SF}^{\text{*}}$+SF+DS+K for evaluating on the KITTI 2015 dataset, where superscript `*' means unsupervised pre-training on the corresponding dataset. 
Table~\ref{table:compare_SOTA} compares our proposed models with existing stereo matching networks. On the KITTI 2015 benchmark,  our proposed \RBDispNetMOC achieves state-of-the-art accuracy among low-cost dense stereo depth estimation methods. Our \RBDispNetMC is 37\%  faster  and  8\%  more accurate  than  FADNet and our \RBDispNetMOC  is  5\% faster and 14\% more accurate than FADNet. Compared with models run at less than 10 FPS, our \RBDispNetMOC is $2\times$ faster than iResNet-i2\cite{iResNet} and $11\times$ faster than PSMNet \cite{PSMNet}, while achieving similar accuracy. On the Scene Flow dataset, our \RBDispNetMC has similar accuracy as DeepPrunner-Fast~\cite{DeepPrunner}, but is $2\times$ faster. Our \RBDispNetMOC model also reaches state-of-the-art accuracy: it has similar performance with DeepPruner-Best \cite{DeepPrunner} and GA-Net \cite{GANet}, but with significant ~$4\times$ and ~$64\times$ speedup, respectively.

\begin{table}[tb!]
\vspace{6pt}
\caption{Quantitative comparison of models on Scene Flow and KITTI 2015 dataset. D1-bg, D1-fg, and D1-all are the percentages of outliers (EPE $\geq 3$ or $\geq 5\%$) averaged over background pixels, foreground pixels, and all ground truth pixels. Runetimes are evaluated on the same platform with RTX 2080Ti GPU.  $\dagger$ indicates the runtime is taken from the original paper, as the code is not available.}

\label{table:compare_SOTA}
\begin{center}
\setlength\tabcolsep{3 pt}

\begin{tabular}{|c|c|c|c|c|c|}
\cline{1-6}
\multirow{2}{*}{Model} & Scene Flow & \multicolumn{3}{c|}{KITTI 2015} & \multirow{2}{*}{Time (ms)}    \\ \cline{2-5}
 & EPE & D1-bg & D1-fg & D1-all &    \\ 
\hline\hline
\multicolumn{6}{|c|}{Time $\leq$ 100 ms} \\ \hline

MADNet~\cite{MADNet}  & -& 3.75\%	&9.20\%	&4.66\%& 14 \\ %2080ti
DispNetC~\cite{SceneflowAndDispNetC} & 1.68 & 4.32\% & 4.41\% & 4.34\% &  20  \\  %2080ti
RTSNet~\cite{RTSNet}  & -& 2.86\%	&6.19\%	&3.41\%& \hspace{3.8pt}20$\dagger$ \\%2080ti
Fast DS-CS~\cite{FastDSCS}  &  2.01& 2.83\%	&4.31\%	&3.08\%& 20 \\ %2080ti
FADNet~\cite{FADNet} & \textbf{0.83} & 2.68\% & \textbf{3.50\%} & 2.82\% & 44  \\ % %2080ti
StereoNet~\cite{StereoNet} & 1.10 & 4.30\% & 7.45\% & 4.83\% & 49   \\ %2080ti
ESMNet~\cite{ESMNet}  & 0.84& 2.57\%	&4.86\%	&2.95\%&  \hspace{3.8pt}67$^\dagger$ \\ 
DeepPruner-Fast~\cite{DeepPrunner} & 0.97& 2.32\%	&3.91\%	&2.59\%& 74 \\ %2080ti
\RBDispNetMC (ours) & \comment{1.03}0.95 & 2.29\% & 4.17\% & 2.60\% & 28  \\ %2080ti
\textbf{\RBDispNetMOC (ours)} & \comment{0.95}\textbf{0.84} & \textbf{2.15\%} & 3.74\% & \textbf{2.42\%} & 42   \\ %2080ti
\hline\hline
\multicolumn{6}{|c|}{Time $>$ 100 ms} \\ \hline
HSM~\cite{HSM}   & -& 1.80\%	&3.85\%	&2.14\%& \textbf{135} \\ 
iResNet-i2~\cite{iResNet} & - &2.25\% &\textbf{3.40\%} &2.44\% & 137 \\ %2080ti
DeepPruner-Best~\cite{DeepPrunner} & 0.86 & 1.87\% & 3.56\%& 2.15\% & 178\\  %2080ti
PSMNet\cite{PSMNet} & 1.09 & 1.86\% & 4.62\% & 2.32\% & 520   \\  %2080ti

GC-Net~\cite{GCNet} & 2.5 & 2.21\% & 6.16\% & 2.87\% & 1030   \\ % %2080ti
GA-Net~\cite{GANet} & \textbf{0.84} & \textbf{1.48\%} & 3.46\% & \textbf{1.81\%} & 2700   \\ \hline % %2.7s on 2080ti

\end{tabular}
\end{center}
\vspace{-0.2cm}
\end{table}

Fig.~\ref{fig:Ablation_Scene Flow} further compares qualitatively the estimated disparity on the Scene Flow test set. \RBDispNetMOC is able to resolve more fine details. The estimated rough occlusion masks from \RBDispNetMOC at scale level $s=2$ are also shown in the last column of Fig.~\ref{fig:Ablation_Scene Flow}. Fig.~\ref{fig:Comparison_KITTI} shows the comparison of the result on the KITTI 2015 benchmark. Both our proposed models, \RBDispNetMC and \RBDispNetMOC, produce higher quality disparity maps when compared with DispNetC \cite{SceneflowAndDispNetC}. They reaches comparable accuracy as PSMNet \cite{PSMNet}, which runs at much smaller frame rate, validating the effectiveness of our proposed improvements.

\section{CONCLUSIONS}

In this paper, we propose \RBDispNetMC, an efficient stereo matching network that efficiently calculates multi-scale cost volume and enables accurate and fast stereo matching. We show that additional occlusion modeling, as is done in our \mbox{\RBDispNetMOC}, further improves the disparity estimation accuracy. Combined with unsupervised pre-training and dataset scheduling strategies, our proposed models achieve the state-of-the-art results compared to other efficient stereo matching networks. We expect our findings to be general and applicable for designing and training more accurate efficient stereo matching networks in the future. 

\newpage
\bibliographystyle{IEEEtran}
\bibliography{ref}

\end{document}